\documentclass[conference]{IEEEtran}
\usepackage[a4paper, right=0.6in, top=1in, bottom=1.21in, left=0.5in]{geometry}
\IEEEoverridecommandlockouts
\usepackage{cite}
\usepackage{amsmath,amssymb,amsfonts}
\usepackage{algorithmic}
\usepackage{graphicx}
\usepackage{textcomp}
\usepackage{xcolor}
\usepackage{booktabs}

\usepackage{tikz}
\usepackage{pgfplots}
\usepackage{pgfplotstable}
\pgfplotsset{compat=1.7}
\usepackage{subcaption}

\def\BibTeX{{\rm B\kern-.05em{\sc i\kern-.025em b}\kern-.08em
    T\kern-.1667em\lower.7ex\hbox{E}\kern-.125emX}}

\begin{document}

\title{Machine Learning-Based Malicious Vehicle Detection for Security Threats and Attacks in Vehicle Ad-hoc Network (VANET) Communications}

\author{\IEEEauthorblockN{\textsuperscript{1}Thanh Nguyen Canh and \textsuperscript{2}Xiem HoangVan$^*$}
\IEEEauthorblockA{
\textit{Department of Robotics Engineering} \\
\textit{University of Engineering and Technology, Vietnam National University, Hanoi, Vietnam}\\
canhthanh@vnu.edu.vn\textsuperscript{1}, xiemhoang@vnu.edu.vn\textsuperscript{2} \\
$^*$\textit{Corresponding authors}
}}

\maketitle

\begin{abstract}
With the rapid growth of Vehicle Ad-hoc Network (VANET) as a promising technology for efficient and reliable communication among vehicles and infrastructure, the security and integrity of VANET communications has become a critical concern. One of the significant threats to VANET is the presence of blackhole attacks, where malicious nodes disrupt the network's functionality and compromise data confidentiality, integrity, and availability. In this paper, we propose a machine learning-based approach for blackhole detection in VANET. To achieve this task, we first create a comprehensive dataset comprising normal and malicious traffic flows. Afterward, we study and define a promising set of features to discriminate the blackhole attacks. Finally, we evaluate various machine learning algorithms, including Gradient Boosting, Random Forest, Support Vector Machines, k-Nearest Neighbors, Gaussian Naive Bayes, and Logistic Regression. Experimental results demonstrate the effectiveness of these algorithms in distinguishing between normal and malicious nodes. Our findings also highlight the potential of machine learning based approach in enhancing the security of VANET by detecting and mitigating blackhole attacks.
\end{abstract}


\section{Introduction}

Vehicle Ad-hoc network (VANET) have emerged as a promising technology to enable efficient and reliable communication among vehicles and infrastructure. Besides, the combination of Intelligent Transport System (ITS) and vans is one of the advantage technologies for the current transportation system to reduce traffic accidents \cite{alrehan2019machine}. With the increasing integration of vehicles with advanced technologies, the security and integrity of VANET communications have become a critical concern. The special characteristics of VANET, such as high mobility, dynamic network topology, etc., compromise the quality of service of the application \cite{kadam2019secure}. In Addition, malicious activities and attacks from nodes within the network pose significant threats, jeopardizing the safety and effectiveness of VANET.

\begin{figure}
    \centering
    \includegraphics[width=0.49\textwidth]{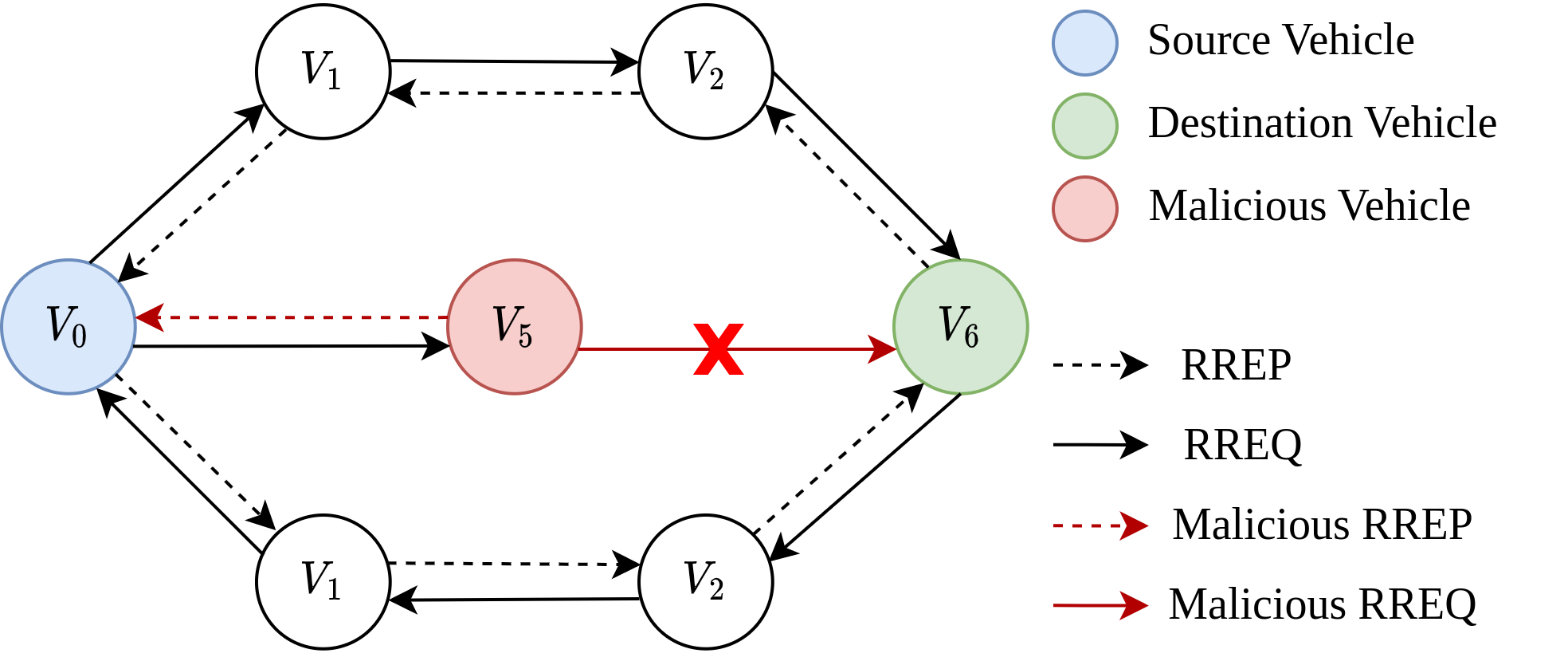}
    \caption{Illustration of a Blackhole Attack in VANET Communications.}
    \label{fig:blackhole}
\end{figure}

Due to the ephemeral nature of VANET, the vehicles engaged in message exchange within the network are susceptible to being targeted by many attacks such as jamming, denial of service (DoS), and especially distributed denial of service (DDoS) attacks \cite{kadam2021machine}. One particular type of attack that can disrupt VANET communications is the presence of malicious nodes, which can compromise the network's functionality with data confidentiality, integrity, and availability. The blackhole attack is very known and dangerous in DoS attacks on the availability in VANET. By disturbing the network layer, which is used for routing in VANET, it has an impact on a user's availability \cite{safdar2022comparison}. An attacker node changes the normal behavior of the routing protocol, and the victim node assumes that it has a valid route to transmit packets to the destination. For VANET, it occurs when a vehicle falsely broadcasts the shortest part to the intended destination of the message. A source vehicle broadcasts a Route Request Packet (RREQ) to all of its nearby vehicles during the route-finding process, and an attacker vehicle interferes with the RREQ and Route Reply Packet (RREP) to drops all the packets and does not forward them to the destination vehicle as shown in Fig. \ref{fig:blackhole}. 

\begin{figure*}
    \centering
    \includegraphics[width=0.83\textwidth]{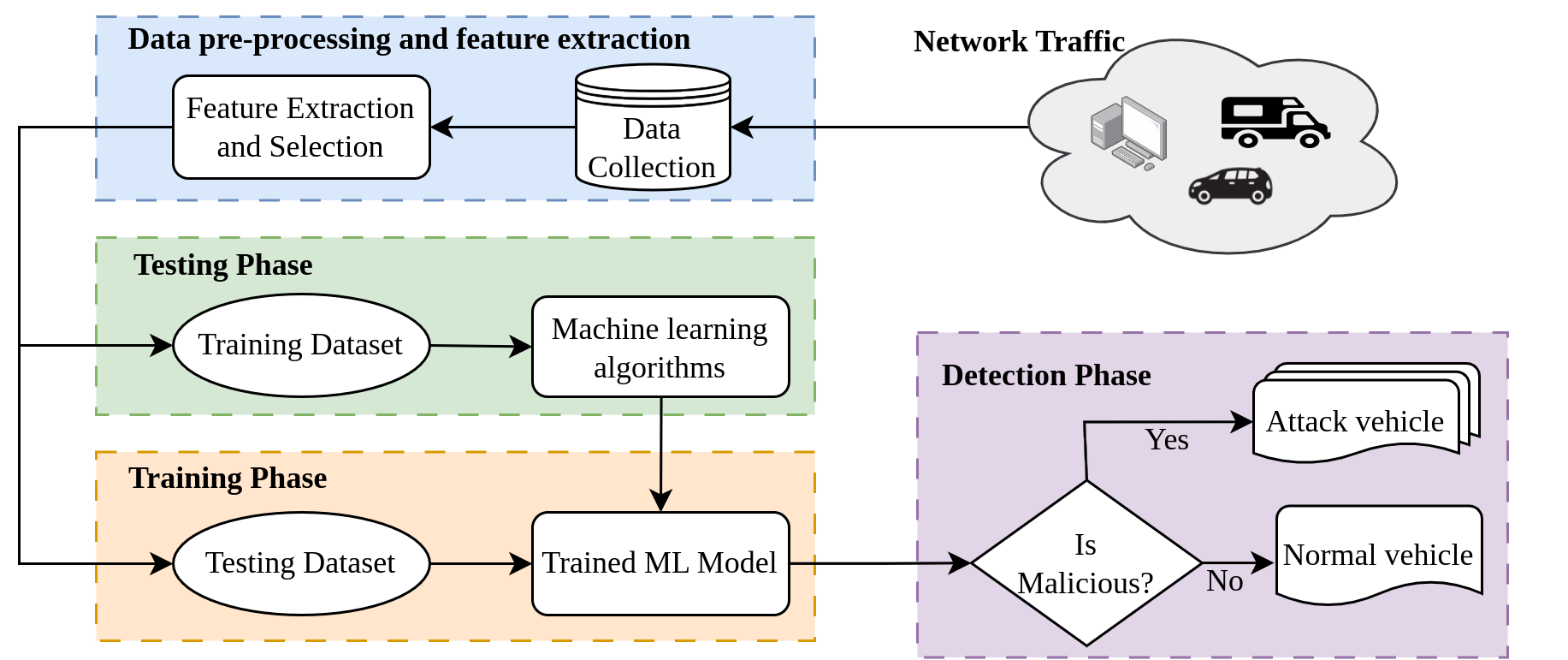}
    \caption{Malicious Vehicle Detection Process Pipeline.}
    \label{fig:overview}
\end{figure*}

In light of these challenges and threats, it is important to develop effective countermeasures to detect and mitigate blackhole attacks and ensure the secure and reliable operation of VANET. This creates conditions that motivate different groups of researchers to evaluate different scenarios \cite{acharya2021dual, lachdhaf2017detection, ahmed2018securing, la2014security, kadam2021machine}. Machine learning recently offers a powerful approach to enhancing model accuracy, especially in scenarios where data structures may dynamically change and large, complex datasets are acquired from specific environments \cite{sugiyama2012machine}. With machine learning algorithms, models can learn from the data, uncover hidden patterns, and adapt to the evolving nature of the dataset. This adaptive nature of machine learning enables it to handle the intricacies and complexities inherent in VANET environments. For example, Zhao et al. \cite{zhao2016svm} implemented a method that uses a greedy forwarding algorithm based on support vector machines (SVM) to enhance both the reliability and communication efficiency in vehicular environments. In a similar study, the works of \cite{singh2019machine, abdan2022machine} demonstrated the potential of using machine learning in wormhole attacks detection. In their next study \cite{montenegro2020detection}, these authors showed that machine learning techniques can effectively utilize the receiver power coherency metric to detect misbehaving nodes in VANET. In \cite{so2018integrating, grover2011machine}, results highlight the effectiveness of k-nearest neighbors (k-NN), support vector machines SVM, and Random Forest (RF) in effectively distinguishing malicious nodes from normal ones, emphasizing their potential as robust and reliable classification techniques in the context of VANET security.

Although the works of \cite{ acharya2021dual, singh2019machine, abdan2022machine, zhao2016svm, so2018integrating, grover2011machine} demonstrated the potential of using machine learning in the classification of malicious detection, they implemented in wormhole attacks \cite{singh2019machine, abdan2022machine} and mobile ad-hoc network (MANET) \cite{abdan2022machine}, the positive predictive value of classification is not high \cite{zhao2016svm, so2018integrating, grover2011machine} and  lack of analysis about features of data \cite{acharya2021dual}. As a matter of fact,  the performance of this approach is rigorous and dependent on the features of the data.

The main objective of this study is  to develop and evaluate a new attach detection method based on a set of discriminate features. Sufficient data encompassing normal and malicious traffic flows has been gathered in order to train and evaluate various machine learning and statistical models, enabling a comparative analysis of their performance. Concretely, we proposed 4 features (source address, destination address, source port, and destination port) for without regard to the parameters on the transmission line for early detection of a blackhole attack. By focusing on these specific attributes, we develop a detection mechanism that can identify the presence of blackhole attacks at an early stage, allowing for timely and effective countermeasures. We used Gradient Boosting (GB), Random Forest (RF), Support Vector Machine (SVM), k-Nearest Neighbors (KNN), Gaussian Naive Bayes (GNB), and Logistic Regression (LR) to evaluated and compared the effectiveness of each classifier against a dataset generated using NS-3 simulator. 

The structure of this paper is as follows. Section \ref{sec:methodology} provides a comprehensive overview of the proposed system and the employed machine learning models. In this section, we discuss the system design and provide insights into the chosen methodologies. In Section \ref{sec:results}, we present the results obtained from our experiments and evaluations, shedding light on the performance and effectiveness of the proposed system. Finally, in Section \ref{sec:conclusion}, we conclude the paper, summarizing the key findings, highlighting the contributions made, and discussing potential avenues for future research in the field.

\section{Methodology and Materials}
\label{sec:methodology}

\begin{table*}
\centering
\caption{Simulation parameters} \label{tab:parameter}
\begin{tabular}{@{}|l|l|@{}}
\toprule
\textbf{Parameter}        & \textbf{Specifications} \\ 
\midrule
Network Simulator & NS 3.29 \\ 
Routing Protocol & Ad-hoc On-Demand Vector Routing (AODV) \\ 
Total number of vehicles & 10 - 65 \\
Total number of malicious vehicles & 1 - 10 \\
Data rate & 600 - 1800 Kbps \\
Number of packets & 7 - 70 \\
Wave packet size & 1024 - 1800 bytes \\
\bottomrule
\end{tabular}
\end{table*}

This section presents the methodology and materials used in our study, outlining the process of estimating malicious vehicles in VANET. The methodology encompasses data collection, feature extraction, and selection, as well as the application of machine learning algorithms for malicious node detection as shown in Fig. \ref{fig:overview}.

\subsection{Data collection}

To train and evaluate our detection models, we conducted extensive data collection involving both normal and malicious traffic flows in VANET. We utilized the NS-3 simulator to simulate these traffic scenarios, allowing us to generate realistic and diverse datasets. Firstly, we utilized the AODV protocol to classify the incoming traffic flow, distinguishing between normal and potentially malicious vehicles based on their communication patterns and behavior. This classification step played a crucial role in enabling the effective detection of malicious nodes within the VANET. Secondly, the AODV protocol facilitated the identification and detection of malicious vehicles. By monitoring the routing and communication activities within the network, we were able to analyze the behavior of each vehicle and identify any signs of malicious intent or anomalous activities. Lastly, our implementation of the AODV protocol aimed to prevent blackhole attacks within the VANET. By continuously monitoring the network and detecting any indications of a blackhole attack, we were able to take proactive measures to mitigate the impact of such attacks, ensuring the integrity and reliability of the VANET communication. To ensure the robustness and reliability of our evaluation, we ran the simulation multiple times, each with varying conditions. These variations included parameters such as the number of normal vehicles, malicious vehicles, data transmission rate, packet size, number of packets, and the initial position of vehicles. By considering a wide range of scenarios, we were able to gather comprehensive data and thoroughly assess the performance of our detection models. Table \ref{tab:parameter} provides an overview of the specific simulation parameters that were varied during the experiments, highlighting the range of conditions under which our detection models were evaluated.

\subsection{Feature extraction and selection}

In the machine learning approach, the selection of informative, discriminating, and independent features plays a crucial role in the effectiveness of classification algorithms. To extract meaningful data from our simulation, we leverage the flow-monitor tool available in NS-3. This tool records comprehensive information about the traffic flow, providing valuable insights for performance analysis \cite{carneiro2009flowmonitor, acharya2021dual}. Considering our objective of detecting malicious behavior as early as possible, while minimizing reliance on detailed traffic information, we carefully choose four independent characteristics from a total of 17 stored features. These characteristics include the outgoing address, destination address, source port, and destination port. By focusing on these specific attributes, we aim to capture essential information related to communication patterns and network behavior without overwhelming the model with unnecessary details. The decision to select these specific features is based on their relevance and potential discriminative power in distinguishing normal and malicious traffic. While the dataset offers a wide range of features, such as source IP address, source, and destination port, time of the first received packet, time of last received packet, delay sum, jitter sum, last delay, lost packets, and throughput, we strategically narrow down our feature set to those that are most likely to contribute to the detection of malicious media. By choosing independent features that capture key aspects of the traffic flow, we aim to strike a balance between the effectiveness of the classification algorithms and the efficiency of feature representation. This approach allows us to develop a robust and efficient detection model that can identify and differentiate between normal and malicious traffic patterns in VANET communication.

\subsection{Machine learning algorithms for malicious node detection}
To classify and detect malicious nodes, we propose to evaluate various machine learning algorithms and the extracted features. The algorithms used in our study include Gradient Boosting (GB), Random Forest (RF), Support Vector Machine (SVM), k-Nearest Neighbors (KNN), Gaussian Naive Bayes (GNB), and Logistic Regression (LR).

\subsubsection{Gradient Boosting (GB)}
The Gradient Boosting algorithm \cite{friedman2002stochastic} is a powerful ensemble learning technique used for classification and regression tasks. It iteratively builds an ensemble of weak learners, typically decision trees, to improve the model's predictive accuracy. By optimizing a loss function through the addition of weak learners, Gradient Boosting effectively learns to make accurate predictions.

The ensemble's prediction at iteration $t$ is given by $F(X)$, which is the sum of the predictions of all the weak learners up to that iteration:

\begin{equation}
    F(X) = \sum_{t=1}^{T} \gamma \cdot h_t(X)
\end{equation}
where $T$ represents the total number of iterations, $\gamma$ is the learning rate, controlling the contribution of each weak learner and $X$ represents the feature matrix.

In this study, the Gradient Boosting Classifier was implemented with specific parameter settings to ensure consistent and reliable results. The chosen settings were a number of estimators of 50, a learning rate of 0.1, a maximum depth of 10, and a random state of 0.

\subsubsection{Random Forest (RF)}

In this paper, we also utilized the Random Forest algorithm to address the challenges of detecting and mitigating blackhole attacks in VANET. Random Forest is an ensemble learning method that combines multiple decision trees to make predictions. Each tree is trained on a random subset of the data and features, and the final prediction is determined by aggregating the predictions of all individual trees.

The Random Forest algorithm \cite{belgiu2016random} provides several advantages in the context of VANET security. It is robust against overfitting, as the randomness introduced in the training process reduces the risk of capturing noise or irrelevant patterns in the data. Additionally, it can handle high-dimensional feature spaces and non-linear relationships effectively. The prediction of the Random Forest algorithm is obtained through a majority voting mechanism. For classification tasks, each tree in the forest independently assigns a class label to a given input instance. The final prediction is then determined by the class label that receives the most votes from all trees. Mathematically, the majority voting can be represented as follows:

\begin{equation}
    f(x) = argmax_c \sum_{i = 1}^{N_{trees}} I(y_i = c)
\end{equation}
where $f(x)$ is the predicted class label, $N_{trees}$ is the total number of trees in the Random Forest, $y_i$ is the class label predicted by $i^{th}$ tree and $c$ represents the class labels.

\subsubsection{Support Vector Machine (SVM)}

SVM \cite{hearst1998support} is a supervised learning algorithm that seeks to find an optimal hyperplane in a high-dimensional feature space, which maximally separates different classes of data points. SVM is particularly effective in handling non-linear classification tasks through the use of kernel functions. The SVM algorithm aims to find the decision boundary that maximizes the margin between the support vectors, which are the data points closest to the decision boundary. These support vectors play a crucial role in defining the decision boundary and are used to classify new, unseen instances. Mathematically, the decision function of an SVM can be represented as:

\begin{equation}\label{eq:svc}
    f(x) = sign \Big ( \sum_{i = 1}^{N_{sv}} y_i \alpha_i K(x_i, x) +b \Big )
\end{equation}
where $f(x)$ is the predicted class label for input instance $x$, $N_{sv}$ is the number of support vectors, $y_i$ is the class label of the $i^{th}$ support vector, $\alpha_i$ represents the corresponding support vector's dual coefficient, $K(x_i, x)$ is the kernel function that measures the similarity between $x_i$ and $x$, and $b$ is the bias term

SVM allows us to handle complex and non-linear decision boundaries by applying different types of kernel functions, such as linear, polynomial, or radial basis function. In this paper, we employed the SVM algorithm with the radial basis function (rbf) kernel for the detection and mitigation of blackhole attacks in VANET. The RBF kernel is particularly suitable for capturing complex and non-linear relationships in the data. Equation \ref{eq:svc} can be represented as:
\begin{equation}
    f(x) = sign \Big ( \sum_{i = 1}^{N_{sv}} y_i \alpha_i \exp ||x_i, x||^2 +b \Big )
\end{equation}
where $\gamma$ is the gamma value controlling the influence of training examples.
\subsubsection{k-Nearest Neighbors (KNN)}
The k-NN algorithm \cite{peterson2009k}is a non-parametric and instance-based classifier that determines the class label of a given instance by examining the class labels of its k nearest neighbors in the feature space. It operates on the assumption that instances with similar feature values tend to belong to the same class.

The k-NN algorithm uses a distance metric to measure the similarity between instances. The most commonly used distance metric is the Euclidean distance, which calculates the straight-line distance between two instances in the feature space. The decision function of the k-NN algorithm can be represented as:

\begin{equation}
    f(x) = majority \ vote (y_1, y_2, \cdots, y_k)
\end{equation}
where $f_x$ represents the predicted class label for the input instance $x$, $y_1, y_2, \cdots, y_k$ are the class labels of the k nearest neighbors of $x$, and the majority vote determines the class label that appears most frequently among the k nearest neighbors.

In our implementation, we set the value of k to 5, which means that the algorithm considers the class labels of the five nearest neighbors to make predictions. This choice strikes a balance between considering sufficient neighbors for robust classification and avoiding overfitting to noise or outliers. By adjusting the value of k, we can control the smoothness of the decision boundary.

\subsubsection{Gaussian Naive Bayes (GNB)}

Naive Bayes \cite{pedregosa2011scikit} is a probabilistic classifier that applies Bayes' theorem with the assumption of independence between features. Gaussian Naive Bayes specifically assumes that the features follow a Gaussian distribution. The algorithm calculates the probability of an instance belonging to a particular class based on the conditional probabilities of its features given that class. The decision function of Gaussian Naive Bayes can be represented as:

\begin{equation}
    f(x) = argmax_c P(C = c) \prod_{i = 1}^n P(X_i = x_i | C = c)
\end{equation}
where $f_x$ represents the predicted class label for the input instance $x$, $C$ is the class variable, $c$ represents a specific class, $P(C = c)$ is the prior probability of class $c$, $X_i$ is the $i^{th}$ feature variable, $x_i$ represents the $i^{th}$ features value of the input instance, and is the $P(X_i = x_i | C = c)$ conditional probability of feature $X_i$ taking the value $x_i$ given class $c$.

By leveraging the probabilistic nature of Gaussian Naive Bayes, we can effectively capture the likelihood of an instance belonging to a certain class based on its feature values. The algorithm's assumption of feature independence allows for computational efficiency and simplicity, making it well-suited for our blackhole attack detection task.
\subsubsection{Logistic Regression (LR)}

The logistic regression \cite{shah2020comparative} model utilizes the sigmoid function, also known as the logistic function, to calculate this probability, considering the linear combination of feature values and their respective coefficients. The sigmoid function, denoted as $\sigma(z)$, transforms the linear combination ($z$) into a value ranging from 0 to 1, representing the probability of the traffic flow being a blackhole attack The equation for the sigmoid function is as follows:

\begin{equation}
    \sigma(z) = \frac{1}{1 + exp^{-z}}
\end{equation}

By comparing the predicted probability to a threshold, typically set at 0.5, we can classify a traffic flow as either normal or indicative of a malicious vehicle.

\section{Experiment Results} \label{sec:results}

In our proposed study, we gathered a dataset comprising 2000 records, with 500 records representing the malicious vehicles and 1500 records representing the normal vehicles in VANET. To ensure a robust evaluation, we divided the dataset into a training set, which consisted of 60$\%$ of the data selected randomly, and a test set containing the remaining 40$\%$. We employed the popular scikit-learn machine learning library in Python to develop our machine learning models. The implementation of the models was carried out using the Kaggle Notebook platform, which provided a conducive environment for efficient development and testing. By leveraging the capabilities of scikit-learn and utilizing Kaggle Notebook's resources, we successfully applied various supervised learning algorithms to assess the effectiveness of our proposed method. These algorithms enabled us to thoroughly evaluate the performance of our approach in detecting and classifying malicious vehicles in VANET.

\subsection{Performance metrics}

To evaluate the performance of the classification algorithms in VANET, we considered popular key metrics such as accuracy, sensitivity, positive predictive value (PPV), and negative predictive value (NPV). Accuracy measures the overall correctness of the predictions, while sensitivity gauges the algorithm's ability to detect the presence of malicious behavior. PPV reflects the proportion of positive classifications, indicating the likelihood of accurate positive predictions, while NPV represents the proportion of negative classifications, indicating the likelihood of accurate negative predictions. These metrics collectively provide a comprehensive assessment of the algorithms' performance and their effectiveness in accurately identifying and classifying malicious behavior in VANET. However, in the case of imbalanced datasets, where the number of records in each class varies significantly, relying solely on accuracy can be misleading. Therefore, we also focus on other performance metrics such as F1-score and ROC AUC score to provide a more robust evaluation of the algorithms' performance.

\begin{equation}
    Accuracy = \frac{TP + TN}{TP + FP + TN + FN}
\end{equation}

\begin{equation}
    Sensitivity = \frac{TP}{TP + FN}
\end{equation}

\begin{equation}
    PPV = \frac{TP}{TP + FP}
\end{equation}

\begin{equation}
    NPV = \frac{TN}{TN + FN}
\end{equation}

\begin{equation}
    F1-score = \frac{2*TP}{2*TP + FP+ FN}
\end{equation}
where TP is True Positives (all true positive), TN is True Negatives, FP is False Positives and FN is False Negatives; PPV is the positive predictive
value and NPV is the negative predictive value

\subsection{Results and discussions}

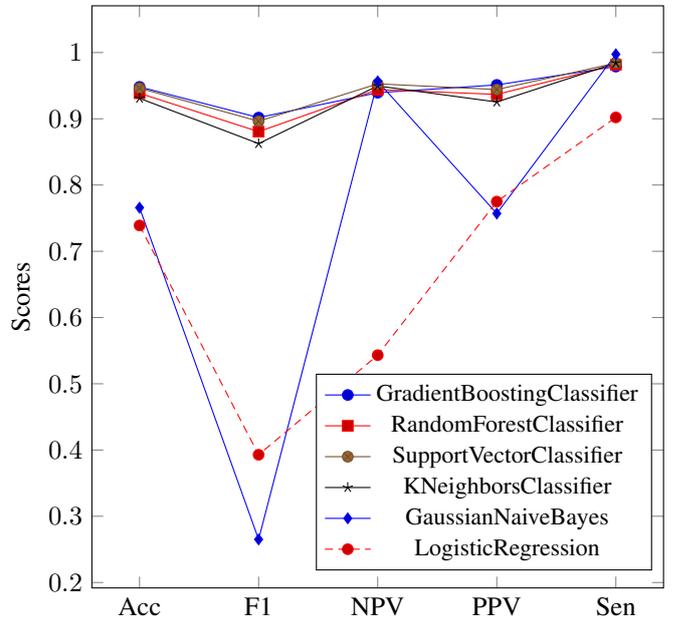
\begin{figure}[!ht]
    \centering
\pgfplotstableread[col sep=comma,]{testdata.csv}\datatable
\pgfplotstabletranspose[colnames from=Method]\TransposedData\datatable
\begin{tikzpicture} 
\begin{axis}[
    width=0.5\textwidth,
    height=9.3cm,
    xtick=data,
    xticklabels from table={\TransposedData}{colnames},
    x tick label style={font=\normalsize},
    legend style={font=\small, at={(0.98,0.03)},anchor=south east},
    ylabel={Scores}]

 \pgfplotsinvokeforeach{1,...,6}{
  \addplot table[x expr=\coordindex,y index=#1] {\TransposedData};
  \pgfmathtruncatemacro{\tmpI}{#1-1}
  \pgfplotstablegetelem{\tmpI}{Method}\of\datatable
  \addlegendentryexpanded{\pgfplotsretval}
}
\end{axis}
\end{tikzpicture}
    \caption{Performance comparison of 6 different machine learning algorithms. GB algorithm achieved the highest performance (accuracy 94.81\% and sensitivity 97.88\%). Legend: Accuracy(Acc), F1-score (F1), Negative Predictive Value (NPV), Positive Predictive Value (PPV), Sensitivity (Sen).}
    \label{fig:result1}
\end{figure}

\begin{figure*}[!ht]
    \centering
\begin{tikzpicture}
\begin{axis}[
    width=\textwidth,
    height= 8cm,
    xlabel={False Positive Rate (FPR)},
    ylabel={True Positive Rate (TPR)},
    xmin= -0.05, xmax=1,
    ymin= 0, ymax=1.1,
    xtick={0,.2,.4,.6,.8,1},
    ytick={0,.2,.4,.6,.8,1},
    legend style={font=\small, at={(0.98,0.03)},anchor=south east},
]

\addplot[mark=o, red!100, tension=0.03 ] 
coordinates {(0,0)(0.02116402,0.86713287)(1,1)}; 
\addlegendentry{GradientBoostingClassifier, ROC AUC Score = 0.923}

\addplot[mark=o, blue!100, tension=0.03 ] 
coordinates {(0,0)(0.01851852,0.82517483)(1,1)}; 
\addlegendentry{RandomForestClassifier, ROC AUC Score = 0.903}

\addplot[mark=o,  orange!100, tension=0.03 ] 
coordinates {(0,0)(0.01587302,0.84615385)(1,1)}; 
\addlegendentry{SupportVectorClassifier, ROC AUC Score = 0.915}

\addplot[mark=o, cyan!100, tension=0.03 ] 
coordinates {(0,0)(0.01587302,0.79020979)(1,1)}; 
\addlegendentry{KNeighborsClassifier, ROC AUC Score = 0.887}

\addplot[mark=o, violet!100, tension=0.03 ] 
coordinates {(0,0)(0.0026455,0.15384615)(1,1)}; 
\addlegendentry{GaussianNaiveBayes, ROC AUC Score = 0.576}

\addplot[mark=o, green!100, tension=0.2 ] 
coordinates {(0,0)(0.0978836,0.30769231)(1,1)}; 
\addlegendentry{LogisticRegression, ROC AUC Score = 0.605}
\addplot[black,dashed] coordinates{(0,0) (0,1)};
\addplot[black,dashed] coordinates{(0,1) (1,1)};
\addplot[black,dashed] coordinates{(0,0) (1,1)};
\end{axis}
\end{tikzpicture}
    \caption{ROC AUC Score Comparision of 6 Different Machine Learning Algorithms.}
    \label{fig:rovauc}
\end{figure*}
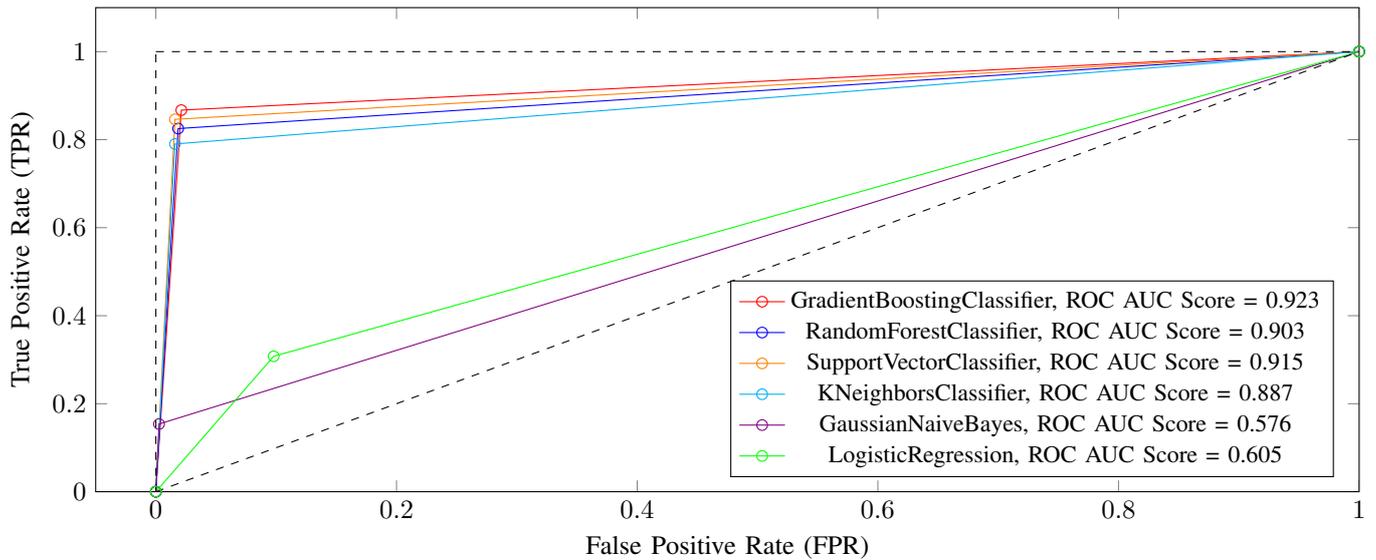

The experiment conducted in this study aims to evaluate the classification performance of six different machine learning methods using various metrics. Figure \ref{fig:result1} illustrates the performance of these classifiers, while Figure \ref{fig:rovauc} presents the ROC-AUC scores. Among the tested classifiers, Logistic Regression and Gaussian Naive Bayes showed lower performance compared to others. When considering all the metrics together, Support Vector Machine and Gradient Boosting classifier demonstrated better and more consistent performance across different scenarios. However, there is a trade-off observed between accuracy, sensitivity, f1 score, positive predictive value, and negative predictive value for these two algorithms. Considering all the factors, it was found that the Gradient Boosting algorithm achieved the highest overall performance. It obtained an overall accuracy of 94.81\%, overall sensitivity of 97.88\%, overall f1-score of 90.18\%, overall NPV of 93.94\%, overall PPV of 95.11\%, and an overall AUC score of 92.3\%.

\section{Conclusion} \label{sec:conclusion}
In this paper, we have presented a comprehensive study on blackhole detection in VANET using machine learning techniques. We proposed a set of discriminate features including source address, destination address, source port, and destination port, which proved to be effective in early detection of blackhole attacks. We evaluated several machine learning algorithms, including Gradient Boosting, Random Forest, Support Vector Machines, k-Nearest Neighbors, Gaussian Naive Bayes, and Logistic Regression, to classify normal and malicious nodes. The experimental results demonstrated the superiority of Gradient Boosting and Random Forest in isolating blackhole nodes, followed by Support Vector Machines and k-Nearest Neighbors. Although Gaussian Naive Bayes and Logistic Regression showed lower performance compared to other classifiers, they still provided valuable insights into the detection process. Overall, our study contributes to the field of VANET security by showcasing the potential of machine learning in detecting and mitigating blackhole attacks. Further research can focus on refining the feature set and exploring advanced machine learning techniques to improve the accuracy and real-time implementation of blackhole detection in VANET.

\bibliographystyle{IEEEtran}
\bibliography{ref}
\end{document}